\documentclass[11pt]{article}
\usepackage{graphicx} % Required for inserting images
\usepackage{amsmath}
\usepackage{lipsum}
\usepackage{pdfcomment}

\usepackage{booktabs, tabularx, makecell}
\usepackage{tikz}
\usetikzlibrary{arrows.meta, positioning, shapes.geometric, fit}
\usepackage{caption}
\usepackage{amsmath}
\usetikzlibrary{positioning}
\usetikzlibrary{arrows.meta, positioning}

\usepackage[table,xcdraw]{xcolor} % Enable table colors and HTML color codes
\usepackage{subcaption}
\usepackage[utf8]{inputenc}
\usepackage{longtable}
\usepackage{graphicx}
\usepackage{float}
\usepackage{booktabs}
\usepackage{float} 
\usepackage{geometry}
\usepackage{lineno}
\usepackage{authblk}
\usepackage{rotating}
\usepackage{booktabs}
\usepackage{threeparttable}
\usepackage{geometry} % Optional: for full page layout control
\usepackage{lscape}
\usepackage{siunitx}  % For aligning numbers on the decimal point
\usepackage{adjustbox} % To adjust table width if needed
\usepackage{pgfplots}
\usepackage{caption}
\usepgfplotslibrary{groupplots}
\usepackage{ragged2e}

\usepackage[numbers,sort&compress]{natbib}

\usepackage{hyperref}
\hypersetup{
  colorlinks=true,
  linkcolor=blue,       % 章节跳转链接（如目录、交叉引用）
  citecolor=blue,       % 文献引用
  urlcolor=blue         % 网页链接
}

\usepackage{cite}

\usepackage{array}
\usepackage{multirow}

\usepackage{amssymb}
\usepackage{pifont}

\usepackage[colorinlistoftodos]{todonotes}

\usepackage{wrapfig}

\usepackage{cite}

\usepackage{amsthm}
\usepackage{titlesec}
\usepackage{titletoc}
\titleclass{\subsubsubsection}{straight}[\subsection]
\newcounter{subsubsubsection}[subsubsection]
\renewcommand\thesubsubsubsection{\thesubsubsection.\arabic{subsubsubsection}}
\titleformat{\subsubsubsection}{\normalfont\normalsize\bfseries}{\thesubsubsubsection}{1em}{}
\titlespacing*{\subsubsubsection}{0pt}{3.25ex plus 1ex minus .2ex}{1.5ex plus .2ex}
\titleformat*{\section}{\fontsize{11}{12}\selectfont\bfseries}
\titleformat*{\subsection}{\fontsize{11}{12}\selectfont\bfseries}
\titleformat*{\subsubsection}{\fontsize{11}{12}\selectfont\bfseries}
\titleformat*{\paragraph}{\fontsize{11}{12}\selectfont\bfseries}
\titleformat*{\subparagraph}{\fontsize{11}{12}\selectfont\bfseries}

\titlecontents{subsubsubsection}
[8em] % adjust this value to fit your needs
{\small}
{\hyperlink{toc.\thecontentslabel}{\thecontentslabel.} }
{}
{\ \titlerule*[.5pc]{.}\contentspage}

\titlespacing*{\subsubsubsection}{0pt}{3.25ex plus 1ex minus .2ex}{1.5ex plus .2ex}

\usepackage{caption}
\usepackage{microtype}

\usepackage{enumitem}
\setlist{itemsep=0em}

\usepackage{fancyhdr} % For custom headers/footers
\usepackage{lastpage} % For total number of pages

\usepackage{listings}
\usepackage{xcolor}

\definecolor{codegreen}{rgb}{0,0.6,0}
\definecolor{codegray}{rgb}{0.5,0.5,0.5}
\definecolor{codepurple}{rgb}{0.58,0,0.82}
\definecolor{backcolour}{rgb}{0.95,0.95,0.92}

\lstdefinestyle{mystyle}{
    backgroundcolor=\color{backcolour},   
    commentstyle=\color{codegreen},
    keywordstyle=\color{magenta},
    numberstyle=\tiny\color{codegray},
    stringstyle=\color{codepurple},
    basicstyle=\ttfamily\footnotesize,
    breakatwhitespace=false,         
    breaklines=true,                 
    captionpos=b,                    
    keepspaces=true,                 
    numbers=left,                    
    numbersep=5pt,                  
    showspaces=false,                
    showstringspaces=false,
    showtabs=false,                  
    tabsize=2
}

\lstdefinestyle{promptstyle}{
    basicstyle=\footnotesize\ttfamily,
    breaklines=true,
    frame=tb,                         % 只有上下横线，没有左右边框
    rulecolor=\color{black!30},       % 线条颜色淡化，不抢眼
    xleftmargin=1.5em,                % 左缩进，增加学术感
    aboveskip=5pt,                    % 进一步压缩外部间距
    belowskip=5pt,
    lineskip=1pt                      % 调节行间距 (Line spacing)
}

\usepackage[most]{tcolorbox}

\newtcolorbox{promptbox}[1][]{
    colback=gray!5,       % 浅灰色背景
    colframe=gray!20,     % 极淡的边框
    arc=1pt,              % 微小圆角
    boxrule=0.5pt,
    left=8pt, right=8pt, top=8pt, bottom=8pt,
    fontupper=\footnotesize\ttfamily, % 使用等宽字体，更显专业
    #1
}

\lstdefinestyle{compactwhite}{
    basicstyle=\footnotesize\ttfamily,      % 小号等宽字体
    breaklines=true,                 % 自动换行
    backgroundcolor=\color{white},   % 纯白背景
    frame=tb,                        % 只显示 Top 和 Bottom 线
    framerule=0.5pt,                 % 线条设细，增加高级感
    rulecolor=\color{black!80},      % 线条颜色（深灰或黑）
    xleftmargin=1em,                 % 紧凑的左边距
    aboveskip=10pt,                  % 调整方框与上方正文间距
    belowskip=10pt,                  % 调整方框与下方标题间距
    lineskip=-0.5pt,                 % 【关键】调这个值：负数会让行间距更紧凑
    columns=flexible                 % 使字符间距更自然
}

\title{Developing an AI Assistant for Knowledge Management and Workforce Training in State DOTs}

\author[1]{Divija Amaram} 
\author[2]{Lu Gao, Ph.D.}
\author[3]{Gowtham Reddy Gudla}
\author[1]{Tejaswini Sanjay Katale}

\affil[1]{Department of Computer Science, University of Houston}
\affil[2]{Department of Civil and Environmental Engineering, University of Houston}
\affil[3]{Engineering Data Science, University of Houston}

\date{}

\begin{document}
\maketitle       
% \linenumbers 

\section*{Abstract}

Effective knowledge management is critical for preserving institutional expertise and improving the efficiency of workforce training in state transportation agencies. Traditional approaches, such as static documentation, classroom-based instruction, and informal mentorship, often lead to fragmented knowledge transfer, inefficiencies, and the gradual loss of expertise as senior engineers retire. Moreover, given the enormous volume of technical manuals, guidelines, and research reports maintained by these agencies, it is increasingly challenging for engineers to locate relevant information quickly and accurately when solving field problems or preparing for training tasks. These limitations hinder timely decision-making and create steep learning curves for new personnel in maintenance and construction operations. To address these challenges, this paper proposes a Retrieval-Augmented Generation (RAG) framework with a multi-agent architecture to support knowledge management and decision making. The system integrates structured document retrieval with real-time, context-aware response generation powered by a large language model (LLM). Unlike conventional single-pass RAG systems, the proposed framework employs multiple specialized agents for retrieval, answer generation, evaluation, and query refinement, which enables iterative improvement and quality control. In addition, the system incorporates an open-weight vision-language model to convert technical figures into semantic textual representations, which allows figure-based knowledge to be indexed and retrieved alongside text. Retrieved text and figure-based context are then provided to an open-weight large language model, which generates the final responses grounded in the retrieved evidence. Moreover, a case study was conducted using over 500 technical and research documents from multiple State Departments of Transportation (DOTs) to assess system performance. The multi-agent RAG system was tested with 100 domain-specific queries covering pavement maintenance and management topics. The results demonstrated Recall@3 of 94.4\%. These results demonstrate the effectiveness of the system in supporting document-based response generation for DOT knowledge management tasks.\\

% keywords can be removed
\noindent \textbf{Keywords:} Knowledge Management, Large Language Models (LLMs), Vision Language Models (VLMs) Retrieval-Augmented Generation (RAG), Pavement Management, AI 

\newpage

\section{Introduction}

Knowledge management (KM) is a set of systematic organizational processes and tools for acquiring, organizing, sustaining, applying, sharing, and maintaining tacit and explicit knowledge so as to enhance organizational and personal performance \citep{su132011387}. In the transportation industry, effective KM is essential for maintaining operational efficiency, ensuring safety, and supporting strategic asset management. This safety linkage is further supported by studies relating pavement performance to crash frequency and severity outcomes, highlighting the potential safety value of timely access to correct pavement knowledge \citep{lebaku2025assessing}. Transportation projects often span long timelines and involve intricate processes that depend on a well-organized body of knowledge, from technical specifications to regulatory requirements and maintenance records \citep{NAP22098}. Knowledge management systems (KMS) provide engineers with quick access to relevant, up-to-date information, which enables them to maintain and repair critical infrastructure that ensures public safety and service continuity \citep{TCRP_KM_2017}. Furthermore, as infrastructure ages and transportation systems expand, the need to retain institutional knowledge becomes critical, particularly when key personnel retire or leave the organization. This also supports asset management by preserving data related to asset life cycles, maintenance schedules, and repair histories, which are instrumental in planning and resource allocation \citep{pantelias2009asset,webb2016schedule}. By utilizing KM, the transportation industry can improve knowledge retention, reduce reliance on individual expertise, and foster a more resilient and adaptable workforce.

Despite its importance, KM in the transportation industry faces several significant challenges. One of the primary obstacles is the reliance on manual processes for capturing and organizing knowledge, which often leads to inconsistencies and inefficiencies in data accessibility and usage \citep{hammer2005}. The transportation sector typically generates vast amounts of data, including technical documents, maintenance logs, safety protocols, and regulatory updates, but much of this information remains stored in unstructured or semi-structured formats that are difficult to retrieve and analyze effectively \citep{torre2018}. Furthermore, the industry's dependence on the expertise of seasoned professionals presents a risk when these experts retire, as critical institutional knowledge may be lost. Hence, the use of artificial intelligence (AI) technologies has been suggested to mitigate the limitations of traditional KM methods in transportation sector \citep{NAP22098}.

Advancements in AI models have significantly enhanced KM across various industries. For example, IBM Watson's decision-support capabilities have been used in healthcare to deliver real-time information, assisting in complex decision-making scenarios where healthcare workers require fast access to critical data \citep{ferrucci2013watson}. Google's DeepMind has shown substantial potential in improving human decision-making in technical environments where immediate access to accurate information is essential \citep{tschang2021}. Adaptive learning systems, such as Microsoft Azure's AI-powered learning platforms, have advanced the personalization of training experiences by dynamically adjusting content based on individual performance metrics. \citet{lim2024} highlight how these platforms allow workers to address skill gaps, which is particularly beneficial in fields where continuous upskilling is required. 

Recently, large language models (LLMs) have been investigated in knowledge retrieval,  particularly in reducing the time workers spend searching for information. \citet{morandini2023} found that chatbots built on these models significantly enhance on-the-job efficiency by quickly retrieving contextually relevant information. However, concerns regarding the accuracy of responses from AI models have been raised  \citep{ozmen2023six}. For example, \citet{jaiswal2023} emphasize ethical considerations, suggesting that while self-learning platforms can support human work, they should ideally complement rather than replace human expertise to avoid ethical and job-related issues.

While recent advancements in LLMs have improved information retrieval efficiency, their reliability remains a concern in high-stakes or knowledge-intensive domains. Consequently, the RAG paradigm has emerged as a promising solution by coupling LLMs with verifiable external knowledge sources \citep{Liu2025NCA_MedRAGSurvey}. RAG overcomes the limitations of LLMs, i.e., outdated information, hallucinations, and absence of domain knowledge. RAG does this by using the strengths of retrieval-based and generation-based approaches. The technique involves applying an information retrieval (IR) system to retrieve relevant information from a knowledge base and feeding it into an LLM to produce contextually correct responses. This combination of the LLM with actual-world information makes it more trustworthy and adaptable for different uses \citep{Hindi2025IEEEAccess}. While previous RAG systems are based on pre-encoded content and embedding vectors, an agentic RAG model enriches data sources with descriptive metadata allowing an agent to dynamically search through diverse data pools. Agentic RAG enhances RAG functionality by using an agent in the construction of proper retrieval actions, mimicking human logical processes for improved accuracy and use in output \citep{Jang2024AURAG}. This adaptability and accuracy in data retrieval, particularly in complex and dynamic information environents, makes agentic RAG unique compared to normal RAG.

Despite recent advances, there remains a gap in deploying LLM-enabled knowledge tools in transportation settings, particularly for State DOT workforce training where domain-grounded accuracy is critical. Prior work often relies on single-pass RAG without explicit quality control, limited multi-step query refinement, or agent specialization. To address this gap, we develop a multi-agent RAG assistant for knowledge management and workforce training that (i) structures retrieval, generation, evaluation, and query refinement into specialized agents, (ii) grounds answers in agency documentation to reduce hallucinations, and (iii) reports transparent performance metrics under realistic workloads. 
In addition, (iv) this study extends conventional text-only RAG by treating figures as retrievable knowledge units rather than auxiliary illustrations, which enables visual evidence in State DOT technical documents to be indexed, retrieved, and directly influence question answering. Unlike prior RAG systems that primarily rely on textual descriptions or ignore figures altogether, the proposed approach allows performance trends, comparative analyses, and expert judgments embedded in figures to actively contribute to retrieval and response generation.  The Retrieved text and figure-based context are
then provided to the Qwen3-4B-Instruct-2507 model, which generates the final responses grounded
in the retrieved evidence.

\section{Literature Review}

\subsection{KM in Infrastructure and Engineering}
KM refers to the systematic processes and tools for acquiring, organizing, sharing, and maintaining both tacit and explicit knowledge within an organization \citep{aviv2021,AlaviLeidner2001MISQ,
DavenportDeLongBeers1998SMR,
GoldMalhotraSegars2001JMIS,
Earl2001JMIS,
LeeChoi2003JMIS,
Bhatt2001JKM,
Bhatt2000JKM,
KankanhalliTanWei2005MISQ,
Wiig1997ESWA,
Nonaka1994OrgSci,
Zack1999CMR}. In engineering-intensive sectors such as transportation and public infrastructure, effective KM is essential for maintaining operational efficiency, ensuring safety, and supporting strategic asset management. Transportation projects often generate vast bodies of knowledge, from technical specifications and standards to maintenance logs and regulatory requirements, which must be well-organized and readily accessible to engineers and decision-makers \citep{gao2013augmented}. In pavement engineering, decision-relevant knowledge is often network-level and context dependent (e.g., influenced by spatial structure), which increases the burden on practitioners to locate and synthesize evidence scattered across heterogeneous technical sources \citep{gao2024considering,gao2021deep}.  By implementing KM systems, infrastructure agencies can provide personnel with quick access to up-to-date information needed to maintain and repair critical assets (e.g. roads, bridges) while upholding public safety and service continuity \citep{NAP22098}. In the transportation domain, several State DOTs have also reported formal efforts to deploy knowledge management systems to support workforce continuity and operational decision-making. For example, \citet{nchrp401} documents knowledge management initiatives across multiple State DOTs, highlighting challenges related to document-heavy workflows, workforce turnover, and decentralized information repositories. These studies emphasize the need for scalable, searchable KM solutions that can operate over large collections of technical reports, manuals, and guidelines typical of DOT environments.

One driving concern for KM in high-complexity, public-sector organizations is the impending retirement of senior experts and the consequent risk of knowledge loss. As infrastructure networks age and seasoned engineers leave, capturing their tacit know-how and lessons learned becomes critical. KM initiatives help convert this tacit institutional knowledge into organizational assets, preserving data on asset life-cycles, maintenance history, and best practices for future generations of engineers. This not only reduces reliance on any single individual's expertise but also fosters a more resilient and adaptable workforce capable of handling complex operations \citep{pantelias2020}. Early efforts in the transportation domain demonstrated the value of formal KM approaches. For example, \citet{hammer2005} explored distributed knowledge management architectures for transportation agencies. \citet{cheung2007systematic} conducted a knowledge audit in a transit organization to identify critical knowledge assets and gaps. These studies highlighted longstanding challenges such as siloed information, fragmented documentation, and inconsistent knowledge capture practices.

Modern AI technologies are now transforming how organizations address these KM challenges. In transportation-focused studies, LLMs have also been used to support analytical tasks such as exploring traffic simulation workflows and cybersecurity strategy reasoning, suggesting that LLM-based assistants can be practical beyond generic Q\&A when grounded in domain context \citep{gao2025exploring}.  A key issue has been the heavy reliance on manual or ad-hoc processes for capturing and organizing knowledge, which leads to inefficiencies in data retrieval and use. Moreover, much of the knowledge in infrastructure agencies resides in unstructured formats (e.g. lengthy reports, PDFs, email communications), which makes it difficult for staff to retrieve relevant information in a timely manner. In addition, infrastructure condition records may contain missing or inconsistent entries that require systematic reconstruction, further reinforcing the need for knowledge tools that help practitioners navigate fragmented and imperfect information \citep{gao2022missing}.  To tackle these problems, researchers have proposed integrating AI into KM systems \citep{GelashviliLuik2025FRAI}. For example, natural language processing and knowledge graph techniques can automatically index and interpret large volumes of engineering documents. \citet{xu2019} demonstrated an ontology-based KM system for highway construction inspection, which structured project knowledge in a machine-readable form to improve information exchange and support decision-making in the field. 

Among emerging AI advances, LLMs and RAG techniques have gained significant attention for KM applications. In closely related AEC settings, LLMs have been explored for compliance-oriented reasoning over code and specification requirements in BIM workflows, indicating that document-grounded LLM pipelines can support rule- and standard-intensive engineering tasks similar to DOT manuals and guidelines \citep{madireddy2025large}.  RAG is an approach that combines an LLM's natural language understanding with an external knowledge base or document repository \citep{lewis2021retrievalaugmentedgenerationknowledgeintensivenlp}. In a RAG pipeline, when a user poses a query, the system first retrieves relevant documents or snippets from the organization's knowledge base, and then the LLM generates a context-aware answer grounded in that retrieved information. By integrating the LLM's output in trusted agency documents, RAG can overcome many limitations of standalone LLMs (such as hallucinations or outdated knowledge) and produce more accurate, domain-specific responses. Researchers are also examining advanced architectures, such as multi-agent or agentic RAG systems \citep{amugongo2025retrieval}. In these systems, specialized agents manage different subtasks. For example, one agent analyzes the query intent, another retrieves focused information from various data sources, and another validates the LLM's answer. These steps aim to mimic human reasoning and enhance the reliability of results. 

\subsection{AI-Enhanced Workforce Training and Upskilling}
Traditionally, agencies have relied on a mix of classroom-based instruction, apprenticeships or mentoring, and static documentation to train new engineers and technicians. These conventional methods, however, often result in fragmented knowledge transfer and inconsistent skill development. In large public agencies (e.g. state DOTs), new hires face steep learning curves due to the sheer complexity of operations and the decades of tacit knowledge held by senior staff. Compounding the challenge, as veteran employees retire, much of their experiential knowledge may leave with them, leading to a “brain drain” that training programs struggle to fill. This problem is especially acute in high-risk, safety-critical fields like infrastructure maintenance, where mistakes by an untrained workforce can have serious consequences. Indeed, recent surveys have found that employees often spend up to 20\% of their work week searching for information or consulting colleagues for help on the job \citep{lee2024house}. The need for more efficient, continuous learning solutions is evident.

AI-based tools are increasingly being explored to augment workforce training and bridge these knowledge gaps \citep{sahraoui2025integrating,Casillo2021ChatbotTrainingEmployees,Limbu2025ChatGPTEmployeeTraining,Gkinko2023ConversationalAIWorkplace,Wilhelm2025ManagersAIConversationalTraining,Naranjo2020VRIndustrialTrainingReview,DiPasquale2024BIMVRIndustrialTraining,Laine2022ITSVRHardSkillsTrainingReview,Tan2025AIAdaptiveLearningPlatformsReview,Prasetya2025AIVocationalEducationReview,Yao2025AIPoweredAdaptiveLearningRAG}
. One promising direction is the use of conversational agents and intelligent tutoring systems powered by large language models. Instead of attending formal classes, trainees can interact with an AI assistant that provides on-demand guidance, answers technical questions, and offers scenario-based problem solving. For example, researchers have developed prototype chatbots using state-of-the-art LLMs fine-tuned on an organization's internal documents and manuals. \citet{lee2024house} report a domain-specific conversational system for engineering document review: by fine-tuning GPT and LLaMA models on thousands of pages of in-house technical specifications, their system could automatically answer engineers' questions about project requirements with a high degree of accuracy. 

Beyond simple Q\&A, AI-driven training platforms can personalize learning pathways and adapt to each employee's progress. Adaptive learning systems use AI to dynamically adjust training content based on individual performance metrics. If a trainee is struggling with a particular procedure or topic, the system can recognize this and provide additional exercises or explanatory material targeting that weakness. Conversely, if an employee demonstrates mastery, the AI can recommend more advanced modules or even suggest practical on-the-job assignments to apply the skill. This level of personalization was traditionally impossible to achieve at scale in large organizations, but AI makes it feasible to deliver a tailored training experience for every worker. \citet{gorowara2024} describe an AI-powered training and reskilling framework that constructs individualized learning paths for employees, using machine learning to select optimal training content and pacing for the digital age workforce.

When integrating generative AI tools (like LLMs) into training environments, researchers stress the importance of maintaining accuracy, trust, and ethics. \citet{shneiderman2022} cautions that heavy reliance on AI recommendations without proper checks can introduce socio-technical risks. In high-risk fields, even a small mistake in training (e.g. a misconstrued safety procedure) can have serious real-world repercussions. Therefore, successful implementations have human experts in the loop: AI-generated answers or training content are validated against authoritative sources or reviewed by instructors, especially in the initial deployment phases. Establishing clear oversight protocols helps build trainees' confidence in the system. Moreover, a theme in recent studies is that AI should augment human mentors, not replace them \citep{jaiswal2023}. While an AI coach can handle routine queries and provide 24/7 support, human trainers are still crucial for teaching nuanced hands-on skills, imparting organizational culture, and providing the empathetic encouragement that algorithms cannot.

Although prior studies on AI-enhanced knowledge management and training platforms have demonstrated the potential to capture, structure, and deliver organizational knowledge more efficiently, there remains limited empirical research on implementing such AI-driven systems within state transportation agencies. This disconnect from the operational realities of DOT environments, where technical documents are extensive, heterogeneous, and continuously updated, limits their practical value. This study addresses these gaps by introducing a multi-agent RAG framework that grounds responses in agency documentation, integrates query refinement and answer evaluation, and demonstrates its effectiveness using a large corpus of State DOT technical and research materials.

\section{Methodology}

This section presents the methodology used to develop the proposed multi-agent RAG framework for knowledge management and workforce training in State DOT environments. The methodology consists of four main components: preprocessing and indexing of heterogeneous document content, including both text and figures; a retrieval-augmented generation pipeline for grounding responses in agency documentation; an agentic workflow that enables structured retrieval, generation, evaluation, and query refinement; and the design of specialized agents that coordinate these tasks. The following subsections describe each component in detail.

\subsection{Figure Preprocessing}

RAG systems rely on text-based representations to perform embedding, similarity search, and retrieval. Consequently, visual elements such as figures, charts, and plots embedded within technical documents cannot be directly processed by standard RAG pipelines. Unlike plain text, figures are not natively represented as semantic tokens and therefore cannot be embedded or retrieved unless they are first converted into a textual form. Without explicit preprocessing, critical technical information conveyed visually such as performance trends, comparative evaluations, threshold values, and practitioner judgments would be excluded from retrieval and reasoning.

State DOT technical documents contain both textual and visual information, where figures often serve as the primary medium for communicating analytical results, model performance comparisons, and expert assessments. In many cases, these insights are only partially described, or not described at all, in the surrounding text. To capture this information in a machine-retrievable form, figures are first extracted from PDF documents and processed using a vision-language model.

\begin{figure}[H]
    \centering
    \includegraphics[width=0.5\linewidth]{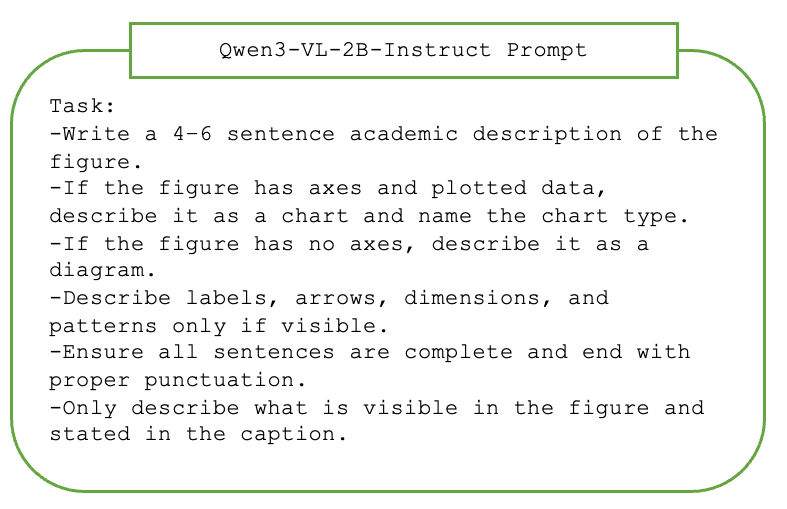}
    \caption{Prompt used for vision-language-based figure description generation.}
    \label{fig:qwen_prompt}
\end{figure}

The objective of figure preprocessing is to translate both the visual content and its associated contextual metadata into structured natural-language representations that preserve the semantic meaning of the original figure. Specifically, the vision-language model is prompted to extract (i) the figure caption  which provides contextual grounding for the visual content, and (ii) a detailed figure description derived from the visual elements themselves. The generated descriptions include information about (a) the type of visualization (e.g., bar chart, line plot, pie chart), (b) the variables shown on each axis, including units and scales, (c) key quantitative relationships and trends such as increases, decreases, or relative differences, (d) categorical comparisons across methods, treatments, or conditions, and (e) qualitative interpretations conveyed through legends, annotations, or practitioner labels. Figure~\ref{fig:qwen_prompt} illustrates the structured prompt used to guide this caption and description generation process. The resulting caption description pairs are treated as first-class knowledge units and indexed alongside textual document chunks in the vector database. By converting figures into semantically rich textual representations that combine contextual labels with visual evidence, the RAG pipeline can apply the same embedding, similarity search, and retrieval mechanisms to both figure-based and text-based information. This preprocessing step enables figure-derived knowledge to be retrieved and used directly during question answering, ensuring that important technical information encoded in figures is not overlooked.

% \subsection{Incorporating Figure-Based Knowledge}
% In addition to text-only queries, the evaluation included questions that relied on information originally presented in figures. As part of the preprocessing pipeline, these figures were converted into structured textual representations and used as retrievable evidence during evaluation. 

% Figures such as pavement performance plots comparing surface texture, skid resistance, roughness, and noise (e.g., Figure~\ref{fig:diamond_grinding}), as well as practitioner opinion charts summarizing model reliability (e.g., Figure~\ref{fig:practitioner_opinion}), were represented as caption-description pairs that capture both contextual labels and detailed quantitative and qualitative interpretations. These figure-derived representations were embedded using the same embedding model as standard text chunks and stored together in the vector database, enabling the RAG system to retrieve and reason over both text-based and figure-based information when answering user queries.

\subsection{RAG}

RAG is a hybrid framework that integrates external knowledge retrieval with generative modeling to enhance the accuracy and trustworthiness of large language models. The process begins with raw documents \( D = \{d_1, d_2, \dots, d_N\} \), such as technical manuals, reports, or articles. Each document \( d_i \) is split into smaller, coherent chunks \( c_{ij} \), where \( i \) denotes the document index and \( j \) represents the chunk index. The resulting set of chunks \( C = \{c_{ij}\} \) ensures that each text chunk is manageable and semantically meaningful.

Each text chunk \( c_{ij} \) is transformed into a high-dimensional vector embedding \( e_{ij} \) using an embedding model \( E \):
\begin{equation}
e_{ij} = E(c_{ij}), \quad e_{ij} \in \mathbb{R}^d
\end{equation}
where,\\
\( E(c_{ij}) \) = Embedding function mapping the chunk \( c_{ij} \) to its vector representation;\\
\( e_{ij} \) = Embedding of the chunk;\\
\( d \) = Dimensionality of the embedding space.

The embeddings for all chunks in a document \( d_i \) are grouped together as:
\begin{equation}
E_i = \{ e_{i1}, e_{i2}, \dots, e_{im} \}
\end{equation}
where \( E_i \) represents the set of embeddings for the chunks of document \( d_i \).

These embeddings for all documents are stored in the Knowledge Base \( V \), which is defined as:
\begin{equation}
V = \{ E_1, E_2, \dots, E_N \} 
\end{equation}
where,\\
\( V \) = The knowledge base containing all chunk embeddings.

When a user submits a query \( q \), it is embedded into a vector representation \( e_q \) using the same embedding model:
\begin{equation}
e_q = E(q), \quad e_q \in \mathbb{R}^d
\label{eq:embedding}
\end{equation}
The query embedding \( e_q \) is compared with the document embeddings \( e_{ij} \) stored in the knowledge base \( V \) to identify the most relevant chunks. Similarity metrics, such as cosine similarity, are used to measure relevance:
\begin{equation}
\text{cosine\_similarity}(e_q, e_{ij}) = \frac{e_q \cdot e_{ij}}{\| e_q \| \| e_{ij} \|}
\label{eq:cosine}
\end{equation}
Alternatively, Euclidean distance (L2) can be used to calculate the nearest vector:
\begin{equation}
\text{distance}(e_q, e_{ij}) = \| e_q - e_{ij} \|
\end{equation}

The retrieved chunks  are concatenated with the query \( q \) to form the context, which is the input for the language model. The context vector is represented as:
\begin{equation}
\text{context} = [q, c_{\text{top-}1}, c_{\text{top-}2}, \dots, c_{\text{top-}k}]
\end{equation}

The language model \( G \) (e.g., Qwen3-4B-Instruct-2507) processes the input context to generate a response \( r \):
\begin{equation}
r = G(\text{context})
\end{equation}
The generative model predicts the response by modeling the probability distribution over the possible sequence of tokens:
\begin{equation}
P(r \mid \text{context}) = \prod_{t=1}^T P(r_t \mid r_{1:t-1}, \text{context})
\end{equation}
Where,\\
\( r \) = The generated response;\\
\( r_t \) = The \( t \)-th token of the response;\\
\( P(r_t \mid r_{1:t-1}, \text{context}) \) = The probability of generating token \( r_t \) given the prior tokens and the context.

The output \( r \) is presented to the user as the final answer, completing the RAG process.

\subsection{Agentic RAG}

The traditional RAG approach performs well for most use cases but has limitations, such as the lack of iterative refinement and multi-step processing, which could enhance answer quality. To address this, we used an agentic RAG system where multiple agents collaborate to improve the response generation process. The system begins with a User Proxy Agent that captures user input, followed by a Retriever Agent that fetches relevant documents. A Generator Agent then formulates the initial response, which is assessed by an Evaluator Agent for accuracy and relevance. If necessary, a Query Refiner Agent further optimizes the query to refine the retrieval process.

\begin{figure}[H]
    \centering
    \includegraphics[width=0.7\linewidth]{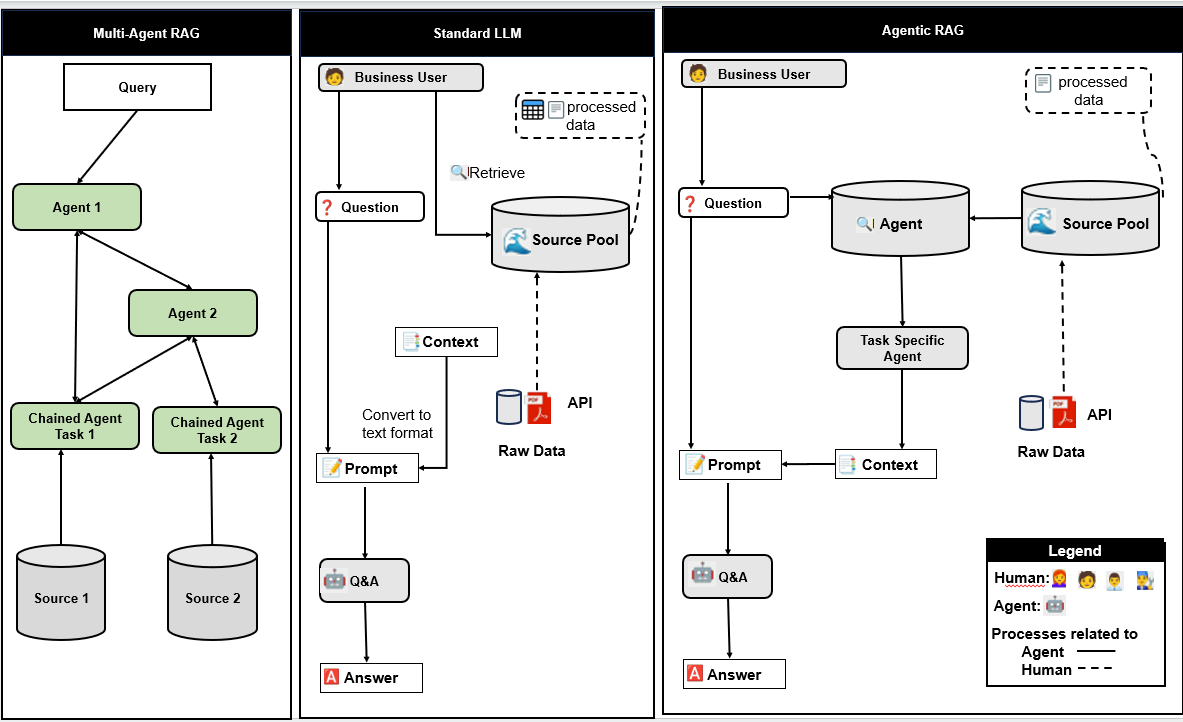}
    \caption{Multi-Agent RAG, Standard LLM, and Agentic RAG architectures for retrieval and response generation. }
    \label{fig:rag_comparison}
\end{figure}

\autoref{fig:rag_comparison} illustrates three RAG approaches: Multi-Agent RAG, Standard LLM, and Agentic RAG, each varying in how queries are processed, information is retrieved, and responses are generated. In this paper, the proposed tool implemented in this work is based on the Multi-Agent RAG approach, where multiple specialized agents collaborate to refine the response generation process. It begins with an initial agent receiving the user query, which is then passed to a retrieval agent to fetch relevant context. Next, a generator agent formulates the response, followed by an evaluation agent that assesses its quality based on predefined criteria. If necessary, a refinement agent further improves the response. In contrast, the Standard LLM approach simply retrieves relevant context and feeds it, along with the query, to a large language model for response generation. The Agentic RAG approach extends standard RAG by introducing task-specific agents that enhance both retrieval and response generation.

The process begins by loading the documents, segmenting them into chunks, generating embeddings, and storing them in the database. The Retriever Agent then retrieves relevant context based on the given query, ensuring that only documents exceeding a predefined relevance threshold are selected. The retrieved context is passed to the Generator Agent, which extracts the answer along with any additional relevant details, such as examples or historical background, if available. The response is then formatted according to the specified structure before being sent to the Evaluator Agent, which assesses it based on predefined criteria, such as whether it is beginner-friendly. If the answer meets the required standards, it is presented as the final output, and the process ends. However, if the response is unsatisfactory, a refinement loop is triggered, where the Query Refiner Agent modifies the query based on feedback from the Evaluator Agent. To prevent infinite looping, query refinement occurs at most k times. If the system is still unable to generate a satisfactory response after k iterations, it displays a predefined message indicating that the answer could not be found. This automated workflow ensures a structured response, with each agent specializing in a specific task to enhance efficiency and accuracy.

\subsection{Agent Design}

In this research, we utilized the AutoGen framework to build the multi-agent system and facilitate communication between agents. AutoGen, developed by Microsoft Research, is an advanced framework designed for constructing multi-agent AI systems that interact, collaborate, and solve complex problems using LLMs. It provides a structured approach to creating AI agents capable of autonomous communication, task planning, and execution. 

The query-answering process begins with a User Proxy Agent, followed by a set of Assistant Agents, including the Retriever Agent, which is specifically instructed to obtain context using the retrieve\_contexts function. Next, the Generator Agent extracts the answer and formats it as required before passing it to the Evaluator Agent for assessment. If necessary, the Query Refiner Agent further refines the query based on feedback. In the following sections, we will provide a detailed discussion on the prompts used for each agent.

\paragraph{User Proxy Agent}
The User Proxy Agent is the initial agent in the multi-agent RAG system. When user input is received, it is first processed by the User Proxy Agent before being passed to subsequent agents. Its primary role is to receive human input and relay it to downstream agents without modification or additional processing. By acting as a seamless intermediary, it ensures a smooth workflow throughout the system. \autoref{fig:user_agent} illustrates the prompt for the User Proxy Agent. As shown, its functionality is clearly defined as to accept user queries and relay them to downstream agents without modification.

\begin{figure}[ht]
\begin{lstlisting}[style=compactwhite]
System Prompt: UserProxyAgent

[Role]
You are the user's representative in the multi-agent system.

[Responsibilities]
- Accept and understand the user's queries.
- Clearly communicate these queries to other agents in the system.
\end{lstlisting}
\caption{System prompt for the \texttt{UserProxyAgent}.}
\label{fig:user_agent}
\end{figure}

\paragraph{Assistant Agent}
In the AutoGen framework, the assistant agent is a specialized component designed to handle specific tasks within a multi-agent system. It can be programmed with defined functions and behaviors, making it highly adaptable for various use cases. The assistant agent interacts with other agents, processes instructions, and executes designated tasks. It can be configured using prompts, function-calling capabilities, and other customization options. In this implementation, assistant agents were responsible for query processing, document retrieval, response generation, and evaluation. Within the Multi-Agent RAG approach, four distinct assistant agents were created and configured to optimize the system's workflow. The retriever agent called the retrieval function to fetch relevant context from the document database. The generator agent extracted relevant information from the retrieved context, formulated a response, and structured it in a predefined format. The evaluator agent assessed the generated response to ensure quality and relevance. Finally, the query refiner agent refined the initial query based on feedback from the evaluator and, if necessary, retriggered the retrieval process to improve the response.

\paragraph{Retriever Agent}

The Retriever Agent is responsible for fetching relevant context from the database based on the given query. Its primary function is to call the retrieve\_contexts function, which performs a similarity-based search to identify and retrieve the most relevant context. Once retrieved, the context is passed to the subsequent agents in the pipeline, specifically the Generator Agent, ensuring a smooth and efficient workflow. \autoref{fig:retriever_agent_instruction} illustrates the prompt for the Retriever Agent, which is explicitly instructed to call the retrieve\_contexts function as its sole responsibility. After retrieving the context, it is passed to the downstream generator agent, which is responsible for generating the response.

\begin{figure}[ht]
\begin{lstlisting}[style=compactwhite]
System Prompt: RetrieverAgent

[Core Responsibility]
Call the function 'retrieve_contexts' to fetch relevant contexts from the database.

[Operational Constraints]
- Do NOT provide responses from internal knowledge.
- Only execute 'retrieve_contexts' using the provided query.
\end{lstlisting}
\caption{System prompt for the RetrieverAgent}
\label{fig:retriever_agent_instruction}
\end{figure}

\paragraph{Generator Agent}

The Generator Agent is responsible for structuring responses in a clear and organized format. It follows specific instructions on formatting the context retrieved by the Retriever Agent and is explicitly guided to provide beginner-friendly responses. \autoref{fig:generator_agent_instruction} illustrates the prompt for the Generator Agent, which outlines its key responsibilities. It must extract the most relevant answer from the retrieved context, ensuring it is concise and directly addresses the user's query. If the context includes relevant examples, trends, or comparisons, the agent summarizes them. Any background information related to the query topic is also included. Also, the agent formulates follow-up questions based on the query topic to enhance user engagement. It is responsible for citing the sources used in its response and is strictly instructed to generate answers solely from the provided context without incorporating external knowledge. The agent maintains a simple and clear tone, and if the retrieved context does not provide sufficient details, it explicitly states that the available information is insufficient to answer the query.

\begin{figure}[ht]
\begin{lstlisting}[style=compactwhite]
System Prompt: GeneratorAgent

[Output Structure]
1. Direct Answer: State the most relevant answer concisely.
2. Examples: Summarize datasets/examples with proper citations.
3. Historical Background: Describe development factors and trends.
4. Follow-Up: Suggest 2-3 questions for deeper exploration.

[Operational Guidelines]
- Use provided context only; do not add outside information.
- If info is missing, use "insufficient information" disclaimer.
- Avoid jargon; maintain a beginner-friendly tone.
- MANDATORY: Always include citations for examples and background.
\end{lstlisting}
\caption{System prompt for the GeneratorAgent}
\label{fig:generator_agent_instruction}
\end{figure}

After extracting the response from the context and formatting it according to the given instructions, the response is then forwarded to the evaluator agent.

\paragraph{Evaluator Agent}

The Evaluator Agent is responsible for assessing the response generated by the Generator Agent based on clarity, relevance, and completeness. It also ensures that the answer is beginner-friendly. If the response meets the defined evaluation criteria, the agent marks it as satisfactory. Otherwise, it deems the response unsatisfactory, provides feedback, and forwards it to the Query Refiner Agent for further refinement. \autoref{fig:evaluator_agent} illustrates the prompt for the Evaluator Agent, clearly outlining the evaluation criteria and specifying the next step based on whether the response is deemed satisfactory or unsatisfactory. Once the evaluator agent completes its assessment, it either determines the response to be satisfactory and returns the final answer or, if deemed unsatisfactory, triggers the query refinement loop by passing the evaluation and feedback to the query refiner agent.

\begin{figure}[ht]
\begin{lstlisting}[style=compactwhite]
System Prompt: EvaluatorAgent
1. Clarity: Is the response beginner-friendly?
2. Relevance: Does it address the query?
3. Completeness: Includes background and follow-ups.

Instruction: Conclude with "Satisfactory" or "Unsatisfactory".
If unsatisfactory, explain required refinements.
\end{lstlisting}
\caption{System prompt for the EvaluatorAgent}
\label{fig:evaluator_agent}
\end{figure}

\paragraph{Query Refiner Agent}
The sole responsibility of the Query Refiner Agent is to refine the query based on feedback from the Evaluator Agent. This agent is essential when the original query is ambiguous or not well-defined. By addressing these challenges, the Query Refiner modifies the query and passes it to the Retriever Agent, triggering the pipeline again. \autoref{fig:query_refiner_agent} illustrates the prompt for the Query Refiner Agent, highlighting its primary function of modifying the query based on the provided feedback.

% \begin{figure}[H]
% \centering
% \begin{tcolorbox}[
%   colback=red!5!white,
%   colframe=red!75!black,
%   title=QueryRefinerAgent,
%   fonttitle=\bfseries\scriptsize,
%   coltitle=white,
%   left=1pt, right=1pt, top=1pt, bottom=1pt,
%   boxsep=0.5pt,
%   arc=0pt
% ]
% \begingroup
% \scriptsize
% \setlength{\baselineskip}{9pt}
% \setlength{\parskip}{0pt}

% You are a query refiner. Your role is to revise the user's query based on feedback from the \texttt{EvaluatorAgent}.

% \begin{enumerate}[itemsep=0pt,parsep=0pt,topsep=1pt,partopsep=0pt]
%     \item Read the feedback from the \texttt{EvaluatorAgent} carefully.
%     \item Identify the original query and pinpoint areas that require improvement.
%     \item Refine the query to make it more specific, clear, and relevant to the user's intent.
%     \item Output only the refined query.
% \end{enumerate}
% \endgroup
% \end{tcolorbox}
% \caption{Instruction for the \texttt{QueryRefinerAgent} in the multi-agent workflow.}
% \label{fig:query_refiner_agent}
% \end{figure}

The retrieved text and figure-based context are then passed to the Qwen3-4B-Instruct-2507 model, which produces the final answers.

\begin{figure}[ht]
\begin{lstlisting}[style=compactwhite]
System Prompt: QueryRefinerAgent
1. Analyze feedback from EvaluatorAgent carefully.
2. Pinpoint areas for improvement in the original query.
3. Refine the query for specificity, clarity, and relevance.
4. Output only the refined query text.
\end{lstlisting}
\caption{System prompt for the QueryRefinerAgent}
\label{fig:query_refiner_agent}
\end{figure}

\section{Case Study}
This section presents a case study to evaluate the proposed system. It is divided into three parts: data description, experimental setup, and results discussion.

\subsection{Data Description}

In this case study, documents collected selected if they (1) were issued or sponsored by state DOTs, federal transportation agencies, or recognized research institutions; (2) focused on pavement maintenance, pavement management systems, materials, rehabilitation strategies, or related pavement decision-making topics; and (3) contained substantive technical content such as design guidance, experimental results, performance models, or implementation practices. Documents that were non-technical, administrative in nature, duplicated across repositories, or lacked sufficient technical depth were excluded. This process resulted in a curated corpus of 521 authoritative documents representative of practical and research-driven pavement management knowledge. Document lengths vary considerably, from technical manuals of just one or two pages to comprehensive reports extending up to 281 pages. Most documents fall within the range of 40 to 130 pages, with a notable portion exceeding 150 pages. The dataset draws on materials from a wide range of state DOTs, national repositories, and international sources. Texas is prominently represented due to the breadth of research and documentation available through TxDOT. In addition, the dataset incorporates data and policy documents from states including California, Kansas, Georgia, North Carolina, Washington, Illinois, South Carolina, Virginia, Oregon, South Dakota, Pennsylvania, Ohio, Missouri, Louisiana, Florida, Utah, Michigan, Maine, Minnesota, Maryland, Tennessee, and New Mexico. 

The topics covered span critical aspects of infrastructure planning and management. Many documents focus on the development and evaluation of Pavement Management Information Systems (PMIS), including the calibration of performance prediction models for asphalt, continuously reinforced concrete, and jointed concrete pavements. Several reports examine revisions to utility curves, updates to decision-tree trigger criteria, and integration of non-destructive testing data into PMIS workflows. Others address comparative analyses between district-level rehabilitation and repair needs versus PMIS-recommended actions. Broader topics include asset management strategy, traffic data analysis, surface drainage modeling with LiDAR, skid resistance evaluation, four-year pavement planning processes, right-of-way acquisition, and the implementation of structural condition indices for network-level pavement evaluation. Some documents also detail specific treatments such as diamond grinding and explore material-related concerns. 

Temporally, the dataset spans over three decades of transportation research and policy development. The oldest documents date back to the 1990s, while the most recent were published in 2024. The majority were released between 2000 and 2023, offering a comprehensive view of both historical evolution and contemporary practices. Many reports cite explicit publication years, such as 2006, 2009, 2010, 2012, 2016, 2017, 2018, 2020, and 2022, and several guidelines and manuals appear in updated editions, with the latest versions from 2023 and 2024.

Based on the collected documents, we extracted a set of recurring technical questions that reflect practitioners' decision needs in pavement management, which focus on treatment service life, cost, application timing, and performance impacts. Table~\ref{tab:extracted_questions_100} summarizes representative questions used for subsequent analysis.

\begin{table}[H]
    \centering
    \footnotesize
    \setlength{\tabcolsep}{8pt}
    \caption{Questions extracted from the curated pavement-management corpus}
    \label{tab:extracted_questions_100}
    
    \begin{tabular}{p{1.5cm} p{12cm}} 
        \hline
        \textbf{No.} & \textbf{Key Investigative Questions} \\
        \hline
        
        % --- 组标题 1 ---
        \multicolumn{2}{l}{\textbf{Surface Treatments: Service Life, Cost, and Application}} \\
        1--7   & What is the expected service life of: chip seal, micro-surfacing, scrub seals, fog seal, seal coat, thin overlays, and light texturing? \\
        8--14  & What is the typical cost associated with these surface treatments? \\
        15--21 & Under what specific weather, traffic, and pavement conditions should these treatments be applied? \\
        \hline
        
        % --- 组标题 2 ---
        \multicolumn{2}{l}{\textbf{Rehabilitation \& Recycling: Strategic Planning}} \\
        22--33 & What is the expected service life for major interventions (e.g., HIR, CIR, FDR, HMA overlays, and various repair/patching methods)? \\
        34--45 & What are the unit costs for these rehabilitation and recycling strategies? \\
        46--57 & What are the critical triggers and conditions for applying these major treatments? \\
        \hline
        
        % --- 组标题 3 ---
        \multicolumn{2}{l}{\textbf{Performance Modeling \& Effectiveness}} \\
        58--64 & What are the observed deterioration rates following surface and HMA overlays? \\
        65--71 & How is the long-term effectiveness of these treatments quantified? \\
        \hline
        
        % --- 组标题 4 ---
        \multicolumn{2}{l}{\textbf{Concrete Pavement \& Technical Triggers}} \\
        72--78 & What are the texturing methods, use cases, and performance impacts for concrete? What are the specific triggers for \textbf{Diamond Grinding}? \\
        \hline
        
        % --- 组标题 5 ---
        \multicolumn{2}{l}{\textbf{System Integration: PMIS Modeling \& Data Governance}} \\
        79--90 & How are performance models calibrated for different pavement types (Asphalt, CRCP, JCP), and how are utility curves or NDT data integrated into workflows? \\
        91--100 & Comparison of district-level needs vs. PMIS recommendations, multi-year planning, network-level metrics (skid resistance), and data quality governance. \\
        \hline
    \end{tabular}
\end{table}

\subsection{Experiment Setup}
Table~\ref{tab:implementation_settings} summarizes the key implementation choices and system parameters used in the case study. 

\begin{table}[H]
\centering
\caption{Implementation details and configuration settings of the proposed multi-agent RAG system}
\label{tab:implementation_settings}
\small % 使用 small 字号，既专业又紧凑
\begin{tabular}{@{}ll@{}} % @{} 去除表格左右两侧多余的边距
\toprule
\textbf{Item} & \textbf{Specification} \\
\midrule
Embedding model & all-MiniLM-L6-v2 \\
Embedding dimensionality & 384 \\
Evaluation metric & Precision@3, Recall@3 \\
Chunk size & 1200 characters \\
Chunk overlap & 0 (no overlap) \\
Vector database & ChromaDB \\
Similarity metric & Cosine similarity \\
Relevance threshold & Not applied (top-$k$ results always returned) \\
Vision-language model & Qwen3-VL-2B-Instruct \\
Response generation model & Qwen3-4B-Instruct-2507 \\
\bottomrule
\end{tabular}
\end{table}

\subsubsection{Preprocessing Figures}

\begin{figure}[H]
    \centering
    
    % --- 第一行左：Fig 4.1 ---
    \begin{subfigure}[t]{0.48\linewidth}
        \centering
        \includegraphics[width=\linewidth]{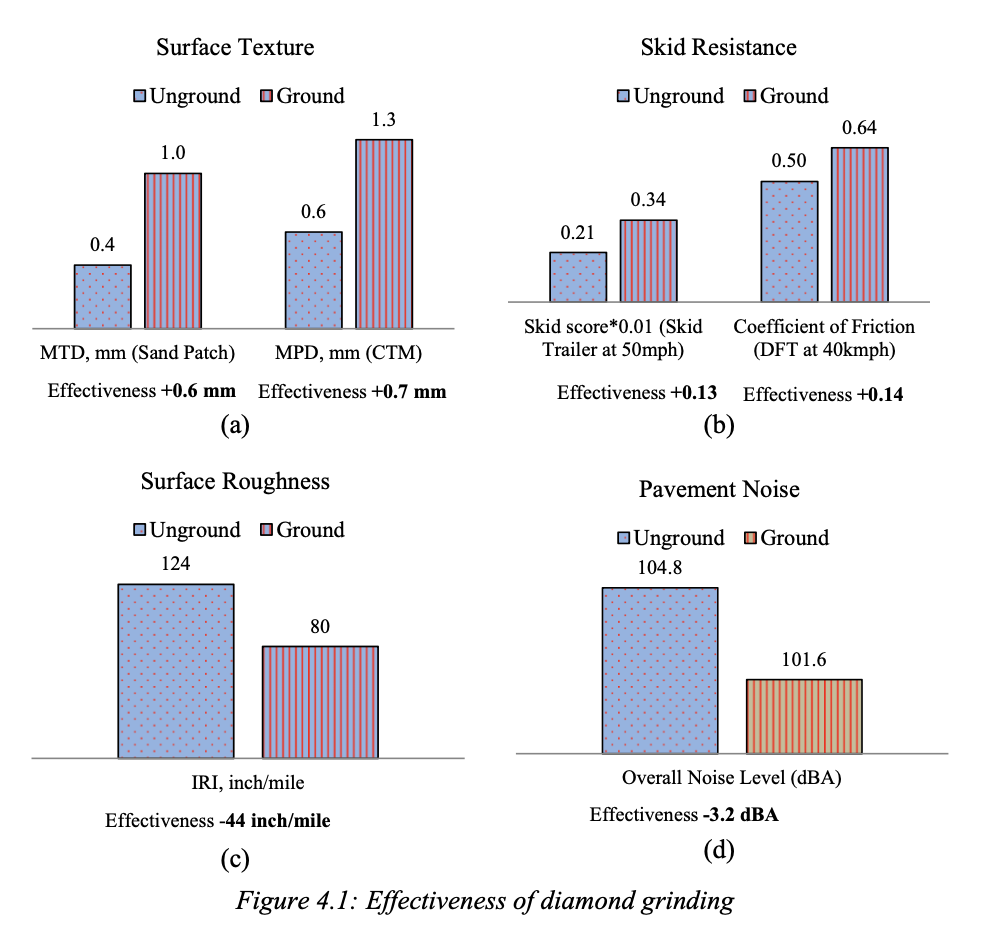}
        \caption{Bar charts adapted from \citet{buddhavarapu2013diamond}.}
        \label{fig:diamond_grinding}
        \vspace{8pt}
        \justifying % 让文字两端对齐，更有宣传册的排版感
        {\footnotesize \textbf{Qwen3-VL-2B-Instruct generated description:} This figure presents four bar charts illustrating the effectiveness of diamond grinding on different pavement surfaces. The first chart, (a), compares surface texture measured by MTD and MPD, showing that the ground surface has higher values than the unground surface for both metrics. The second chart, (b), evaluates skid resistance, indicating that the ground surface provides better performance than the unground surface. The third chart, (c), shows that the unground surface has a higher IRI value than the ground surface, while the fourth chart, (d), demonstrates that the overall noise level is lower on the ground surface compared to the unground surface. Each chart uses distinct colors and patterns to differentiate between the two surface types.\par}
    \end{subfigure}
    \hfill
    % --- 第一行右：Fig 8 ---
    \begin{subfigure}[t]{0.48\linewidth}
        \centering
        \includegraphics[width=\linewidth]{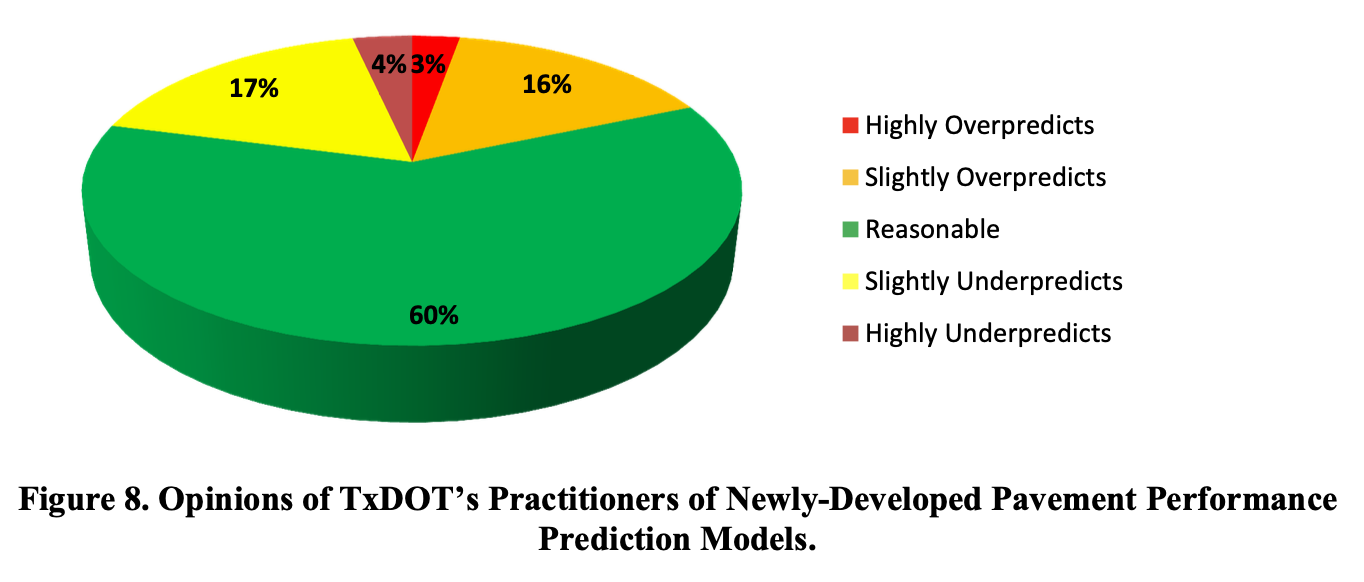}
        \caption{Pie chart adopted adapted from \citet{gharaibeh2012pmis}.}
        \label{fig:practitioner_opinion}
        \vspace{8pt}
        \justifying
        {\footnotesize \textbf{Qwen3-VL-2B-Instruct generated description:} This figure is a pie chart that displays the opinions of TxDOT's practitioners regarding newly-developed pavement performance prediction models. The chart is divided into five segments, each representing a different level of prediction accuracy with corresponding percentages. The largest segment, colored green, accounts for 60\% and represents ``Reasonable'' predictions. The second-largest segment, yellow, represents ``Slightly Underpredicts'' at 17\%. The smallest segments are ``Highly Overpredicts'' (red, 4\%) and ``Slightly Overpredicts'' (orange, 16\%). The legend on the right side of the chart identifies each category by its color and label.\par}
    \end{subfigure}

    % \vspace{2.5em} % 两行之间的垂直间距

    % \vspace{1em}
    \caption{VLM Description of Figures}
    \label{fig:pavement_overview_2x2}
\end{figure}

Figure~\ref{fig:diamond_grinding} shows pavement performance comparisons in terms of surface texture, skid resistance, roughness, and noise, while Figure~\ref{fig:practitioner_opinion} shows practitioner opinions summarizing model reliability. These figures were represented as caption-description pairs that capture both contextual labels and detailed quantitative and qualitative interpretations. The figure-derived representations were embedded using the same embedding model as standard text chunks and stored together in the vector database, which enables the RAG system to retrieve and reason over both text-based and figure-based information when answering user queries.

% \begin{figure}[h]
%     \centering
%     % --- Subfigure 1 ---
%     \begin{subfigure}[b]{0.28\linewidth}
%         \centering
%         \includegraphics[width=\linewidth]{Fig_4.1.png}
%         \caption{Pavement performance comparison for diamond grinding on CRCP, adapted from \citet{buddhavarapu2013diamond}.}
%         \label{fig:diamond_grinding}
%     \end{subfigure}
%     \hfill
%     % --- Subfigure 2 ---
%     \begin{subfigure}[b]{0.48\linewidth}
%         \centering
%         \includegraphics[width=\linewidth]{Fig_8.png}
%         \caption{Practitioner opinion chart on pavement model reliability, adapted from \citet{gharaibeh2012pmis}.}
%         \label{fig:practitioner_opinion}
%     \end{subfigure}
    
%     % --- Main Caption and Label ---
%     \caption{Comparison of pavement performance metrics and practitioner opinions.}
%     \label{fig:pavement_overview}
% \end{figure}

\subsubsection{Evaluation Metrics}

The system was evaluated using 100 domain-specific questions (Table \ref{tab:extracted_questions_100}). For each question, an LLM (Qwen3-4B-Instruct-2507) was prompted to assess every retrieved document and determine whether it was relevant to the query. The LLM-assigned relevance labels were then manually reviewed and verified against ground truth. Retrieval quality was measured using Precision@3 and Recall@3, where Precision@3 measures how many of the top three retrieved documents are relevant, and Recall@3 measures how many of relevant documents are found within the top three results returned by the system. 

% The evaluation questions were designed to reflect common information needs in pavement maintenance and management. They covered topics such as expected service life and cost of pavement treatments (e.g., chip seals, microsurfacing, fog seals, hot-mix asphalt overlays, and full-depth reclamation), appropriate application conditions, treatment timing, and deterioration behavior. This set of questions was chosen to assess the system's ability to retrieve relevant technical information and generate clear, well-structured answers.

\begin{equation}
\text{Recall@3} =
\frac{\text{Number of relevant documents in the top 3 retrieved}}
{\text{Total number of relevant documents}}
\end{equation}

\begin{equation}
\text{Precision@3} =
\frac{\text{Number of relevant documents in the top 3 retrieved}}
{\text{3}}
\end{equation}

\subsection{Results Discussion}

To evaluate the performance of the proposed system, we tested it using 100 domain-specific queries (Table \ref{tab:extracted_questions_100}) and compared the system's responses against manually verified ground truth. Retrieval quality was measured using Precision@3 and Recall@3. The system achieved a Precision@3 of close to 1.0 and a Recall@3 of 94.4\%. A Precision@3 of 1.0 indicates that all documents retrieved within the top three results were relevant, while a Recall@3 of 94.4\% indicates that nearly all relevant evidence was retrieved within the top three results. Together, these results show that the system consistently retrieves accurate and sufficient contextual information for downstream answer generation.

To assess result stability, retrieval metrics were computed across multiple queries and reported as macro-averages. Variability across queries was observed but remained consistent, which indicates stable retrieval behavior across different topics. After retrieval, the top-ranked text chunks and figure-derived descriptions were provided as context to the Qwen3-4B-Instruct-2507 model, which generated the final responses grounded in the retrieved evidence.

\subsubsection{Single-Pass RAG vs. Multi-Agent RAG}

To understand the contribution of the proposed multi-agent design, a study was conducted comparing a single-pass RAG baseline with the multi-agent RAG system. Both systems used the same document corpus, embedding model, and large language model, which ensures that the observed differences are due to the retrieval and reasoning strategy rather than model capacity. Retrieval performance was evaluated using top-$k$ recall and precision. Using the same set of 100 domain-specific evaluation queries, the single-pass RAG system achieved a Recall@5 of 58\%, which indicates that relevant documents were frequently missing even when up to five documents were retrieved. In contrast, the proposed multi-agent RAG system achieved a Recall@3 of 94\%, meaning that relevant evidence was successfully retrieved within the top three results for nearly all queries. The improvement in recall is due to the multi-agent workflow, where retrieval is refined through evaluator feedback rather than executed only once. When an initial response lacks sufficient evidence, the Query Refiner Agent reformulates the query and triggers additional retrieval steps, allowing the system to recover relevant documents that are commonly missed in a single-pass setup. This comparison shows that the multi-agent RAG system finds important documents more reliably than a single-pass RAG by refining queries when key information is missing, which leads to better evidence retrieval for DOT-related questions.

\section{Conclusion}

This paper presented a multi-agent RAG system designed to support knowledge management and workforce training in State DOTs. By separating retrieval, response generation, evaluation, and query refinement into specialized agents, the system is able to produce structured and document-grounded answers rather than relying on single-pass language model outputs. Using a corpus of more than 500 technical and research documents from multiple State DOTs and a set of 100 domain-specific queries related to pavement maintenance and management, the system achieved a Precision@3 of 1.0 and a Recall@3 of 94.4\%, which indicates that all documents retrieved within the top three results were relevant and that nearly all relevant evidence was successfully retrieved within this top-ranked set.

The primary limitation of the current study lies in the scope of the dataset. The proposed system is evaluated using pavement-related documents, which limits its applicability to other transportation domains. As a result, the findings primarily reflect the system's effectiveness within the pavement engineering context and may not fully generalize to broader transportation infrastructure areas such as traffic operations, safety analysis, or bridge management.

This system is designed to operate under low-cost deployment constraints using standard computing infrastructure without specialized hardware or on premise servers. Document ingestion is performed periodically, and figure based information is currently incorporated through manual extraction and description generation rather than fully automated pipelines. While this approach reflects realistic State DOT workflows and avoids the complexity and cost of hard automation, it limits scalability and requires human involvement (e.g, manual screenshot capture of figures). These constraints highlight practical trade offs between cost, automation, and scalability, and motivate future work on incremental automation of figure processing as deployment resources permit.

Future work will focus on expanding the dataset beyond pavement engineering to include a wider range of transportation domains. Incorporating reports related to traffic engineering, roadway safety, asset management, and multimodal transportation systems will enable a more comprehensive evaluation of the framework's generalizability. This expansion will help assess the system's adaptability across diverse transportation applications and support its potential use as a unified knowledge assistant for transportation agencies.

%Bibliography
\bibliographystyle{unsrtnat}
\bibliography{references}

@book{NAP22098,
  author    = "{Transportation Research Board and National Academies of Sciences, Engineering and Medicine}",
  title     = "A Guide to Agency-Wide Knowledge Management for State Departments of Transportation",
  isbn      = "978-0-309-30887-8",
  doi       = "10.17226/22098",
  url       = "https://nap.nationalacademies.org/catalog/22098/a-guide-to-agency-wide-knowledge-management-for-state-departments-of-transportation",
  year      = 2015,
  publisher = "The National Academies Press",
  address   = "Washington, DC"
}

@book{TCRP_KM_2017,
  author    = {{National Academies of Sciences, Engineering, and Medicine}},
  title     = {Knowledge Management Resource to Support Strategic Workforce Development for Transit Agencies},
  year      = {2017},
  publisher = {National Academies Press},
  doi       = {10.17226/24961},
  url       = {https://nap.nationalacademies.org/read/24961/chapter/1}
}

@Article{su132011387,
AUTHOR = {Aviv, Itzhak and Hadar, Irit and Levy, Meira},
TITLE = {Knowledge Management Infrastructure Framework for Enhancing Knowledge-Intensive Business Processes},
JOURNAL = {Sustainability},
VOLUME = {13},
YEAR = {2021},
NUMBER = {20},
ARTICLE-NUMBER = {11387},
URL = {https://www.mdpi.com/2071-1050/13/20/11387},
ISSN = {2071-1050},
DOI = {10.3390/su132011387}
}

@inproceedings{Hammer2005,
    author = {Hammer, Joachim and Langer, Hagen and Timm, Ingo J.},
    title = {Distributed Knowledge Management in the Transportation Domain},
    booktitle = {Proceedings of the 10th International Symposium on Logistics},
    year = {2005},
    pages = {489--491},
    address = {Lisbon}
}

@article{Pantelias2020,
    author = {Pantelias, A. and others},
    title = {Data-Driven Asset Management in Transportation Infrastructure},
    journal = {Journal of Infrastructure Systems},
    year = {2020},
    volume = {26},
    number = {2},
    pages = {45-58}
}

@article{Torre2018,
    author = {Torre-Bastida, Ana Isabel and Del Ser, Javier and Laña, Ibai and Ilardia, Maitena and Bilbao, Miren Nekane and Campos-Cordobés, Sergio},
    title = {Big Data for transportation and mobility: recent advances, trends and challenges},
    journal = {IET Intelligent Transport Systems},
    year = {2018},
    volume = {12},
    number = {8},
    pages = {742--755}
}

@article{Morandini2023,
    author = {Morandini, M. and Lazzarini, G. and Mauri, D.},
    title = {The impact of artificial intelligence on workers' skills: Upskilling and reskilling in organizations},
    journal = {Journal of Organizational Learning},
    year = {2023},
    volume = {29},
    number = {4},
    pages = {650-670},
    doi = {10.1177/15344843231224009}
}

@article{tschang2021,
    author = {Tschang, F. T. and Almirall, E.},
    title = {Artificial intelligence as augmenting automation: Implications for employment},
    journal = {Academy of Management Perspectives},
    year = {2021},
    volume = {35},
    number = {2},
    pages = {286-307},
    doi = {10.5465/amp.2019.0062}
}

@article{lim2024,
    author = {Lim, S. and Lee, K.},
    title = {Rethinking education in the era of artificial intelligence (AI): Towards future workforce competitiveness},
    journal = {Artificial Intelligence in Education},
    publisher = {Springer},
    year = {2024},
    doi = {10.1007/978-981-97-2211-2_7}
}

@article{gorowara2024,
    author = {Gorowara, V. and Kapoor, A. and Wadhwa, P.},
    title = {AI personalizing training and reskilling employees for the digital age},
    journal = {IEEE Xplore},
    year = {2024},
    doi = {10.1109/ICRAI.2024.10497293}
}

@article{shneiderman2022,
    author = {Shneiderman, B.},
    title = {Artificial intelligence: Socio-technical risks and accuracy challenges},
    journal = {AI \& Society},
    year = {2022},
    volume = {37},
    number = {3},
    pages = {567-578}
}

@article{jaiswal2023,
    author = {Jaiswal, N. and Sinha, R. and Kumar, S.},
    title = {Rebooting employees: Upskilling for artificial intelligence in multinational corporations},
    journal = {Artificial Intelligence and Future HRM},
    year = {2023},
    pages = {15-27},
    doi = {10.4324/9781003377085-5}
}

@misc{lewis2021retrievalaugmentedgenerationknowledgeintensivenlp,
      title={Retrieval-Augmented Generation for Knowledge-Intensive NLP Tasks}, 
      author={Patrick Lewis and Ethan Perez and Aleksandra Piktus and Fabio Petroni and Vladimir Karpukhin and Naman Goyal and Heinrich Küttler and Mike Lewis and Wen-tau Yih and Tim Rocktäschel and Sebastian Riedel and Douwe Kiela},
      year={2021},
      eprint={2005.11401},
      archivePrefix={arXiv},
      primaryClass={cs.CL},
      url={https://arxiv.org/abs/2005.11401}, 
}

@article{ozmen2023six,
  title   = {Six human-centered artificial intelligence grand challenges},
  author  = {Ozmen Garibay, Ozlem and Winslow, Brent and Andolina, Salvatore and Antona, Margherita and Bodenschatz, Anja and Coursaris, Constantinos and Falco, Gregory and Fiore, Stephen M. and Garibay, Ivan and Grieman, Keri and Havens, John C. and Jirotka, Marina and Kacorri, Hernisa and Karwowski, Waldemar and Kider, Joe and Konstan, Joseph and Koon, Sean and Lopez-Gonzalez, Monica and Maifeld-Carucci, Iliana and McGregor, Sean and Salvendi, Gavriel and Ben Shneiderman and Stephanidis, Constantine and Strobel, Christina and Ten Holter, Carolyn and Xu, Wei},
  journal = {International Journal of Human--Computer Interaction},
  year    = {2023},
  volume  = {39},
  number  = {3},
  pages   = {391--437},
  publisher = {Taylor \& Francis}
}

@article{amugongo2025retrieval,
  title={Retrieval augmented generation for large language models in healthcare: A systematic review},
  author={Amugongo, Lameck Mbangula and Mascheroni, Pietro and Brooks, Steven and Doering, Stefan and Seidel, Jan},
  journal={PLOS Digital Health},
  volume={4},
  number={6},
  pages={e0000877},
  year={2025},
  publisher={Public Library of Science San Francisco, CA USA}
}

@article{pantelias2009asset,
  title={Asset management data practices for supporting project selection decisions},
  author={Pantelias, Aristeidis and Flintsch, Gerardo W and Bryant Jr, James W and Chen, Chen},
  journal={Public Works Management \& Policy},
  volume={13},
  number={3},
  pages={239--252},
  year={2009},
  publisher={SAGE Publications Sage CA: Los Angeles, CA}
}

@article{ferrucci2013watson,
  title={Watson: beyond jeopardy!},
  author={Ferrucci, David and Levas, Anthony and Bagchi, Sugato and Gondek, David and Mueller, Erik T},
  journal={Artificial Intelligence},
  volume={199},
  pages={93--105},
  year={2013},
  publisher={Elsevier}
}

@article{lee2024house,
  title={In-house knowledge management using a large language model: focusing on technical specification documents review},
  author={Lee, Jooyeup and Jung, Wooyong and Baek, Seungwon},
  journal={Applied Sciences},
  volume={14},
  number={5},
  pages={2096},
  year={2024},
  publisher={MDPI}
}

@article{Hindi2025IEEEAccess,
  author  = {Hindi, Mahd and Mohammed, Linda and Maaz, Ommama and Alwarafy, Abdulmalik},
  title   = {Enhancing the Precision and Interpretability of Retrieval-Augmented Generation (RAG) in Legal Technology: A Survey},
  journal = {IEEE Access},
  year    = {2025},
  volume  = {13},
  pages   = {46171--46189},
  doi     = {10.1109/ACCESS.2025.3550145},
  url     = {https://ieeexplore.ieee.org/document/10921633}
}

@inproceedings{Jang2024AURAG,
  author    = {Jang, Jisoo and Li, Wen{-}Syan},
  title     = {AU-RAG: Agent-based Universal Retrieval Augmented Generation},
  booktitle = {Proceedings of the ACM SIGIR Asia Pacific (SIGIR-AP)},
  year      = {2024},
  pages     = {2--11},
  doi       = {10.1145/3673791.3698416},
  url       = {https://dl.acm.org/doi/pdf/10.1145/3673791.3698416}
}

@article{Liu2025NCA_MedRAGSurvey,
  author  = {Liu, Shuaishuai and others},
  title   = {A survey on retrieval-augmentation generation (RAG) for medicine},
  journal = {Neural Computing and Applications},
  year    = {2025},
  doi     = {10.1007/s00521-025-11666-9},
  url     = {https://link.springer.com/article/10.1007/s00521-025-11666-9}
}

@article{cheung2007systematic,
  title={A systematic approach for knowledge auditing: a case study in transportation sector},
  author={Cheung, Chi Fai and Li, ML and Shek, WY and Lee, Wing Bun and Tsang, TS},
  journal={Journal of knowledge management},
  volume={11},
  number={4},
  pages={140--158},
  year={2007},
  publisher={Emerald Group Publishing Limited}
}

@article{Aviv2021,
author = {Aviv, Itzhak and Hadar, Irit and Levy, Meira},
title = {Knowledge Management Infrastructure Framework for Enhancing Knowledge-Intensive Business Processes},
journal = {Sustainability},
volume = {13},
number = {20},
pages = {11387},
year = {2021},
doi = {10.3390/su132011387}
}

@article{Xu2019,
author = {Xu, Xin and Yuan, Chenxi and Zhang, Yuxi and Cai, Hubo and Abraham, Dulcy M. and Bowman, Mark D.},
title = {Ontology-Based Knowledge Management System for Digital Highway Construction Inspection},
journal = {Transportation Research Record},
volume = {2673},
number = {1},
pages = {52--65},
year = {2019},
doi = {10.1177/0361198118823499}
}

@article{GelashviliLuik2025FRAI,
  author  = {Gelashvili-Luik, Teona and Vihma, Peeter and Pappel, Ingrid},
  title   = {Navigating the AI revolution: challenges and opportunities for integrating emerging technologies into knowledge management systems. Systematic literature review},
  journal = {Frontiers in Artificial Intelligence},
  year    = {2025},
  volume  = {8},
  pages   = {1595930},
  doi     = {10.3389/frai.2025.1595930},
  url     = {https://www.frontiersin.org/articles/10.3389/frai.2025.1595930/full}
}

@techreport{nchrp401,
  title        = {Advances in Transportation Agency Knowledge Management},
  author       = {{National Cooperative Highway Research Program}},
  institution  = {Transportation Research Board},
  number       = {NCHRP Synthesis 401},
  year         = {2010},
  address      = {Washington, DC},
  url          = {https://onlinepubs.trb.org/onlinepubs/nchrp/nchrp_syn_401.pdf},
  note         = {Accessed: 2025-03-17}
}

@techreport{buddhavarapu2013diamond,
  author       = {Buddhavarapu, Prasad and de Fortier Smit, André and Prozzi, Jorge A. and Trevino, Manuel},
  title        = {Evaluation of the Benefits of Diamond Grinding of CRCP: Final Report},
  institution  = {Texas Department of Transportation},
  number       = {FHWA/TX-13/5-9046-01-1},
  year         = {2014},
  address      = {Austin, TX},
  note         = {Performed August 2011--August 2013; published November 2014}
}

@techreport{gharaibeh2012pmis,
  author       = {Gharaibeh, Nasir and Wimsatt, Andrew and Saliminejad, Siamak and Menendez, Jose Rafael and Weissmann, Angela Jannini and Weissmann, Jose and Chang-Albitres, Carlos},
  title        = {Implementation of New Pavement Performance Prediction Models in PMIS: Report},
  institution  = {Texas A\&M Transportation Institute},
  number       = {FHWA/TX-12/5-6386-01-1},
  year         = {2012},
  address      = {College Station, TX},
  note         = {Published November 2012. TxDOT Project 5-6386-01},
  url          = {http://tti.tamu.edu/documents/5-6386-01-1.pdf}
}

@article{gao2013augmented,
  title={An augmented Lagrangian decomposition approach for infrastructure maintenance and rehabilitation decisions under budget uncertainty},
  author={Gao, Lu and Guo, Runhua and Zhang, Zhanmin},
  journal={Structure and Infrastructure Engineering},
  volume={9},
  number={5},
  pages={448--457},
  year={2013},
  publisher={Taylor \& Francis}
}

@article{webb2016schedule,
  title={Schedule compression impact on construction project safety},
  author={Webb, Curt and Gao, Lu and Song, Ling-guang},
  journal={Frontiers of Engineering Management},
  volume={2},
  number={4},
  pages={344--350},
  year={2016}
}

@article{gao2021deep,
  title={A deep learning approach for imbalanced crash data in predicting highway-rail grade crossings accidents},
  author={Gao, Lu and Lu, Pan and Ren, Yihao},
  journal={Reliability Engineering \& System Safety},
  volume={216},
  pages={108019},
  year={2021},
  publisher={Elsevier}
}

@article{gao2022missing,
  title={Missing pavement performance data imputation using graph neural networks},
  author={Gao, Lu and Yu, Ke and Lu, Pan},
  journal={Transportation research record},
  volume={2676},
  number={12},
  pages={409--419},
  year={2022},
  publisher={SAGE Publications Sage CA: Los Angeles, CA}
}

@article{gao2024considering,
  title={Considering the spatial structure of the road network in pavement deterioration modeling},
  author={Gao, Lu and Yu, Ke and Lu, Pan},
  journal={Transportation Research Record},
  volume={2678},
  number={5},
  pages={153--161},
  year={2024},
  publisher={SAGE Publications Sage CA: Los Angeles, CA}
}

@article{madireddy2025large,
  title={Large Language Model-Driven Code Compliance Checking in Building Information Modeling},
  author={Madireddy, Soumya and Gao, Lu and Din, Zia Ud and Kim, Kinam and Senouci, Ahmed and Han, Zhe and Zhang, Yunpeng},
  journal={Electronics},
  volume={14},
  number={11},
  pages={2146},
  year={2025},
  publisher={MDPI}
}

@article{AlaviLeidner2001MISQ,
  title   = {Review: Knowledge Management and Knowledge Management Systems: Conceptual Foundations and Research Issues},
  author  = {Alavi, Maryam and Leidner, Dorothy E.},
  journal = {MIS Quarterly},
  year    = {2001},
  volume  = {25},
  number  = {1},
  pages   = {107--136},
  doi     = {10.2307/3250961}
}

@article{DavenportDeLongBeers1998SMR,
  title   = {Successful Knowledge Management Projects},
  author  = {Davenport, Thomas H. and De Long, David W. and Beers, Michael C.},
  journal = {Sloan Management Review},
  year    = {1998},
  volume  = {39},
  number  = {2},
  pages   = {43--57}
}

@article{GoldMalhotraSegars2001JMIS,
  title   = {Knowledge Management: An Organizational Capabilities Perspective},
  author  = {Gold, Andrew H. and Malhotra, Arvind and Segars, Albert H.},
  journal = {Journal of Management Information Systems},
  year    = {2001},
  volume  = {18},
  number  = {1},
  pages   = {185--214}
}

@article{Earl2001JMIS,
  title   = {Knowledge Management Strategies: Toward a Taxonomy},
  author  = {Earl, Michael J.},
  journal = {Journal of Management Information Systems},
  year    = {2001},
  volume  = {18},
  number  = {1},
  pages   = {215--233}
}

@article{LeeChoi2003JMIS,
  title   = {Knowledge Management Enablers, Processes, and Organizational Performance: An Integrative View and Empirical Examination},
  author  = {Lee, Heeseok and Choi, Byounggu},
  journal = {Journal of Management Information Systems},
  year    = {2003},
  volume  = {20},
  number  = {1},
  pages   = {179--228}
}

@article{Bhatt2001JKM,
  title   = {Knowledge Management in Organizations: Examining the Interaction between Technologies, Techniques, and People},
  author  = {Bhatt, Ganesh D.},
  journal = {Journal of Knowledge Management},
  year    = {2001},
  volume  = {5},
  number  = {1},
  pages   = {68--75},
  doi     = {10.1108/13673270110384419}
}

@article{Bhatt2000JKM,
  title   = {Organizing Knowledge in the Knowledge Development Cycle},
  author  = {Bhatt, Ganesh D.},
  journal = {Journal of Knowledge Management},
  year    = {2000},
  volume  = {4},
  number  = {1},
  pages   = {15--26},
  doi     = {10.1108/13673270010315371}
}

@article{KankanhalliTanWei2005MISQ,
  title   = {Contributing Knowledge to Electronic Knowledge Repositories: An Empirical Investigation},
  author  = {Kankanhalli, Atreyi and Tan, Bernard C. Y. and Wei, Kwok-Kee},
  journal = {MIS Quarterly},
  year    = {2005},
  volume  = {29},
  number  = {1},
  pages   = {113--143},
  doi     = {10.2307/25148670}
}

@article{Wiig1997ESWA,
  title   = {Knowledge Management: Where Did It Come From and Where Will It Go?},
  author  = {Wiig, Karl M.},
  journal = {Expert Systems with Applications},
  year    = {1997},
  volume  = {13},
  number  = {1},
  pages   = {1--14},
  doi     = {10.1016/S0957-4174(97)00018-3}
}

@article{Nonaka1994OrgSci,
  title   = {A Dynamic Theory of Organizational Knowledge Creation},
  author  = {Nonaka, Ikujiro},
  journal = {Organization Science},
  year    = {1994},
  volume  = {5},
  number  = {1},
  pages   = {14--37},
  doi     = {10.1287/orsc.5.1.14}
}

@article{Zack1999CMR,
  title   = {Developing a Knowledge Strategy},
  author  = {Zack, Michael H.},
  journal = {California Management Review},
  year    = {1999},
  volume  = {41},
  number  = {3},
  pages   = {125--146}
}

@inproceedings{gao2025exploring,
  title={Exploring Traffic Simulation and Cybersecurity Strategies Using Large Language Models},
  author={Gao, Lu and Liu, Yongxin and Chen, Hongyun and Liu, Dahai and Zhang, Yunpeng and Sun, Jingran},
  booktitle={2025 IEEE Security and Privacy Workshops (SPW)},
  pages={346--351},
  year={2025},
  organization={IEEE}
}

@article{lebaku2025assessing,
  title={Assessing the Influence of Pavement Performance on Road Safety Through Crash Frequency and Severity Analysis: PKR Lebaku et al.},
  author={Lebaku, Prathyush Kumar Reddy and Gao, Lu and Sun, Jingran and Wang, Xingju and Kang, Xuejian},
  journal={International Journal of Pavement Research and Technology},
  pages={1--22},
  year={2025},
  publisher={Springer Nature Singapore Singapore}
}

@article{sahraoui2025integrating,
  title={Integrating generative ai in bim education: Insights from classroom implementation},
  author={Sahraoui, Islem and Kim, Kinam and Gao, Lu and Din, Zia and Senouci, Ahmed},
  journal={arXiv preprint arXiv:2507.05296},
  year={2025}
}

@inproceedings{Casillo2021ChatbotTrainingEmployees,
  author    = {Mario Casillo and Francesco Colace and Massimo De Santo and Marco Lombardi and Domenico Santaniello},
  title     = {A Chatbot for Training Employees in Industry 4.0},
  booktitle = {Research and Innovation Forum 2020 (RIIFORUM 2020)},
  series    = {Springer Proceedings in Complexity},
  pages     = {397--409},
  year      = {2021},
  publisher = {Springer, Cham},
  doi       = {10.1007/978-3-030-62066-0_30}
}

@inproceedings{Limbu2025ChatGPTEmployeeTraining,
  author    = {Digamber Sendang Limbu and S. Vijayakumar Bharathi},
  title     = {An Empirical Study on the Critical Factors Influencing the Application of ChatGPT in Employee Training Programmes},
  booktitle = {Proceedings of International Conference on Data Analytics and Insights (ICDAI 2024)},
  series    = {Lecture Notes in Networks and Systems},
  volume    = {1234},
  pages     = {277--287},
  year      = {2025},
  publisher = {Springer, Singapore},
  doi       = {10.1007/978-981-96-2329-7_22}
}

@article{Gkinko2023ConversationalAIWorkplace,
  author  = {Lorentsa Gkinko and Amany Elbanna},
  title   = {The appropriation of conversational AI in the workplace: A taxonomy of AI chatbot users},
  journal = {International Journal of Information Management},
  volume  = {69},
  pages   = {102568},
  year    = {2023},
  doi     = {10.1016/j.ijinfomgt.2022.102568}
}

@inproceedings{Wilhelm2025ManagersAIConversationalTraining,
  author    = {Lance T. Wilhelm and Xiaohan Ding and Kirk McInnis Knutsen and Buse Carik and Eugenia H. Rho},
  title     = {How Managers Perceive AI-Assisted Conversational Training for Workplace Communication},
  booktitle = {Proceedings of the ACM International Conference on Conversational User Interfaces (CUI '25)},
  year      = {2025},
  eprint    = {2505.14452},
  archivePrefix = {arXiv},
  primaryClass  = {cs.HC}
}

@article{Naranjo2020VRIndustrialTrainingReview,
  author  = {Jose E. Naranjo and Diego G. Sanchez and Angel Robalino-Lopez and Paola Robalino-Lopez and Andrea Alarcon-Ortiz and Marcelo V. Garcia},
  title   = {A Scoping Review on Virtual Reality-Based Industrial Training},
  journal = {Applied Sciences},
  volume  = {10},
  number  = {22},
  pages   = {8224},
  year    = {2020},
  doi     = {10.3390/app10228224}
}

@article{DiPasquale2024BIMVRIndustrialTraining,
  author  = {Valentina Di Pasquale and Salvatore Digiesi and Ivan Ferretti and Antonio Padovano and Chiara Sammarco and Javier Ernesto Su{\'a}rez Savigne},
  title   = {Enhancing Industrial Operator Training through BIM-Enriched Virtual Reality Scenes},
  journal = {IFAC-PapersOnLine},
  volume  = {58},
  number  = {19},
  pages   = {253--258},
  year    = {2024},
  doi     = {10.1016/j.ifacol.2024.09.183}
}

@article{Laine2022ITSVRHardSkillsTrainingReview,
  author  = {Joakim Laine and Timo Lindqvist and Tiina Korhonen and Kai Hakkarainen},
  title   = {Systematic Review of Intelligent Tutoring Systems for Hard Skills Training in Virtual Reality Environments},
  journal = {International Journal of Technology in Education and Science},
  volume  = {6},
  number  = {2},
  pages   = {178--203},
  year    = {2022},
  doi     = {10.46328/ijtes.348}
}

@article{Tan2025AIAdaptiveLearningPlatformsReview,
  author  = {Le Ying Tan and Shiyu Hu and Darren J. Yeo and Kang Hao Cheong},
  title   = {Artificial intelligence-enabled adaptive learning platforms: A review},
  journal = {Computers and Education: Artificial Intelligence},
  pages   = {100429},
  year    = {2025},
  doi     = {10.1016/j.caeai.2025.100429}
}

@article{Prasetya2025AIVocationalEducationReview,
  author  = {Febri Prasetya and Aprilla Fortuna and Agariadne Dwinggo Samala and Dinda Khaira Latifa and Welli Andriani and Utari Akhir Gusti and Muhammad Raihan and Santiago Criollo-C and Deniz Kaya and Juan Luis Cabanillas Garc{\'i}a},
  title   = {Harnessing artificial intelligence to revolutionize vocational education: Emerging trends, challenges, and contributions to SDGs 2030},
  journal = {Social Sciences \& Humanities Open},
  pages   = {101401},
  year    = {2025},
  doi     = {10.1016/j.ssaho.2025.101401}
}

@article{Yao2025AIPoweredAdaptiveLearningRAG,
  author  = {Yao Yao and Horacio Gonz{\'a}lez-V{\'e}lez},
  title   = {AI-Powered System to Facilitate Personalized Adaptive Learning in Digital Transformation},
  journal = {Applied Sciences},
  volume  = {15},
  number  = {9},
  pages   = {4989},
  year    = {2025},
  doi     = {10.3390/app15094989}
}

\end{document}